\PassOptionsToPackage{table}{xcolor}   
\documentclass[11pt]{article}
\usepackage[]{acl}
\usepackage{times}
\usepackage{latexsym}
\usepackage[T1]{fontenc}
\usepackage[utf8]{inputenc}
\usepackage{comment}
\usepackage{microtype}
\usepackage{graphicx}
\usepackage{booktabs}
\usepackage{amsmath}
\usepackage{amssymb}
\usepackage{xcolor}
\usepackage{url}
\usepackage{adjustbox}
\usepackage[most]{tcolorbox}   
\hypersetup{pdftitle={A Comprehensive Evaluation of Decision Models in Real-World Contact Center Conversational Tasks},pdfauthor={}}

\providecommand{\new}[1]{#1}

\definecolor{rowDecision}{HTML}{E3F2FD}
\definecolor{rowLLM}{HTML}{FFF3E0}

\title{Can Decision Models Replace LLMs in Real-World Contact Center Classification and Judging Tasks?\\\texorpdfstring{%
  \raisebox{-0.2\height}{\includegraphics[height=7mm]{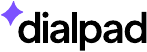}}%
}{}
}

\author{Md Tahmid Rahman Laskar, Xue-Yong Fu, Shashi Bhushan TN\\\{\texttt{tahmid.rahman,xue-yong,sbhushan}\}\texttt{@dialpad.com}\\{Dialpad Inc.}}

\begin{document}
\maketitle

\begin{abstract}
{Contact center} systems often need to make bounded decisions, such as {identifying a caller's intent, retrieving the knowledge-base article that answers a query, assessing agent compliance,} or judging the output of AI models used in production. Decision models return a typed answer and its probability directly, making them a promising cost-effective alternative to generative Large Language Models (LLMs) in such bounded {conversational} tasks. However, their effectiveness in real-world {contact center applications} remains unclear. In this paper, we compare {nine} decision models against {thirteen} generative LLMs in zero-shot settings across eight {contact center} use cases: six classification tasks and two judging tasks. Our experiments demonstrate that {many} decision models can match the performance of frontier LLMs {on contact center classification, and {API-based closed} decision models achieve this at the lowest per-request cost among the API models we evaluate}. We further observe that their confident predictions are more accurate, enabling selective routing in real-world production scenarios.  {Across classification tasks, the main weakness of most decision models is abstention: they rarely choose the no-answer option, which accounts for most of their gap to the strongest LLMs (although Gemini-Decision-1 is the notable exception in this regard).} We also find that the judging quality of decision models 
drops sharply for smaller models, and they are also more prone to position bias. In short, our findings provide practical guidance for selecting decision models based on abstention behavior, confidence, and deployment cost, rather than overall accuracy alone.
\end{abstract}
\section{Introduction}
Large Language Models (LLMs) are increasingly used to analyze business conversations, generate summaries, and power conversational agents~\citep{laskar2023building,fu-etal-2024-tiny,laskar-etal-2025-ai,laskar2026from}. Yet many {contact center} tasks require a bounded decision rather than a generated response. For instance, {identifying a caller's intent, checking whether a human agent followed a required procedure during a call, selecting the knowledge-base article that answers a customer's question, or scoring the output of an AI model} can be considered bounded decision tasks. More specifically, such tasks can only have a set of allowed answers. Applying an LLM to such decision tasks introduces substantial inference cost~\citep{li2026jevjudge}.
\begin{figure*}[!t]
\centering\includegraphics[width=\textwidth]{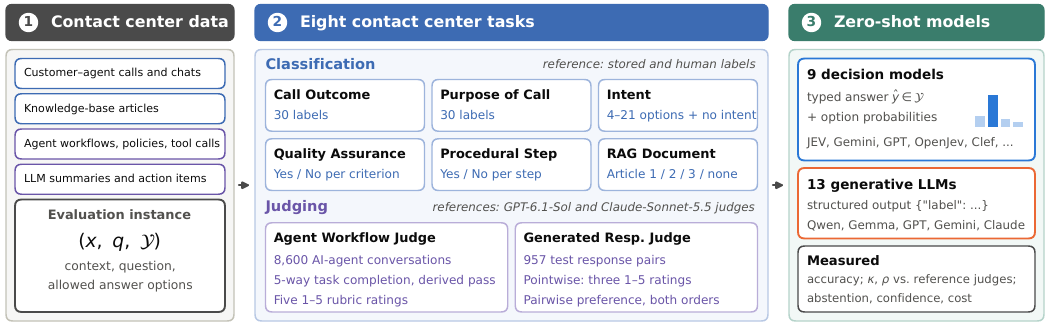}
\caption{Overview of the evaluation. Decision models and generative LLMs answer the same instances zero-shot.}
\label{fig:overview}
\end{figure*}

To tackle this, decision models offer an alternative cost-effective interface. Given a context, a question, and the allowed answer options, they return a typed decision together with a probability for each option~\citep{typesafe2026,clef2026,google2026decisions,openai2026decisions}. Unlike encoder classifiers such as BERT~\citep{devlin2019bert}, {decision models} demonstrate strong performance in zero-shot settings, without task-specific training data. This makes them attractive for {real-world contact center applications}, where training data construction is not trivial since label sets may differ across customers \cite{dong2021semi,parikh2023exploring,pattnaik2025scalable}. 

In this paper, we investigate whether zero-shot decision models can replace generative LLMs for {contact center} decision problems, i.e., tasks that do not necessarily require free-form text generation. We compare {nine} decision models with {thirteen} generative LLMs on six classification and two judging tasks (Figure~\ref{fig:overview}). Our major contributions:
\begin{itemize}\setlength\itemsep{-4pt}
\item We conduct a comprehensive zero-shot evaluation of decision models and generative LLMs on eight real-world {contact center} tasks, covering classification tasks to model judging. 
\item We show that {most} decision models rarely choose the no-answer option, which accounts for most of their gap to the strongest LLMs.
\item We find that judge quality depends on the task and metric: decision models align well with {frontier LLM judges} on pointwise ratings, but poorly on pairwise preference{, where they also show strong position bias}.
\item We further analyze model confidence and find that confident predictions are more accurate.
\end{itemize}

\begin{table}[t]
\centering\setlength{\tabcolsep}{2pt}
\begin{adjustbox}{max width=\columnwidth}
\begin{tabular}{lrl}
\toprule
\textbf{Task} & \textbf{Samples} & \textbf{Answer options} \\
\midrule
Call Outcome & {524} & 30 labels \\
Purpose of Call & {173} & 30 labels \\
Quality Assurance & {187} & Yes / No \\
Procedural Step & {263} & Yes / No \\
Intent & 762 & {4--21 options} \\
RAG Document & {485} & {3 articles or none} \\
\midrule
Agent Workflow Judge & 8,600 & {5-way verdict, 5 ratings} \\
Generated Resp. Judge & {957} & {3 ratings, pairwise preference} \\
\bottomrule
\end{tabular}
\end{adjustbox}
\caption{{Evaluation tasks. Quality Assurance and Procedural Step contain 1,036 and 2,938 scored questions, respectively.}}
\label{tab:tasks}
\end{table}

\section{Related Work}
Zero-shot text classification uses task descriptions to predict labels without target-task training~\citep{yin2019,laskar-etal-2023-systematic}. Decision models extend this setting with typed outputs and native probability distributions~\citep{clef2026,openjev2026}. Recent work studies confidence-based escalation when JEV is used as a judge~\citep{li2026jevjudge} and delegation between decision models and LLM agents~\citep{zhu2026jevagentharness}. However, prior work mostly evaluates decision models on public academic benchmarks, where zero-shot results are exposed to data contamination~\citep{li2026specialized,deusser2026evaluating}. We instead examine them on proprietary real-world {contact center} datasets. 

Another important use case of decision models is to use them as the judge to evaluate different AI models' generated responses \cite{li2026jevjudge}. However, prior work {demonstrates} that smaller, cost-effective LLM judges fail to match human assessments~\citep{fu-etal-2023-large}, require task-specific adaptation~\citep{laskar-etal-2025-improving}, and are prone to position bias~\citep{zheng2023,wang2024,laskar2025judging}. We therefore evaluate decision-model judges on both their agreement with reference judgments and biases in verdicts. 
\section{Tasks and Data}
\label{sec:tasks}
We collect real-world data from Dialpad\footnote{\url{https://www.dialpad.com/}}, covering customer--agent call conversation transcripts, retrieved documents for assisting agents, and AI-generated responses. We organize it into six classification tasks and two judging tasks (see Table~\ref{tab:tasks} and Figure~\ref{fig:overview}). 
Below, we describe each task. 

\subsection{Classification Tasks}
\paragraph{Call outcome and purpose of call.}
Given a customer--agent call transcript, \emph{Call Outcome} asks for the result of the call (e.g., issue resolved, transferred), while \emph{Purpose of Call} asks for the customer's reason for calling (e.g., billing questions). Both tasks require the selection of one label from 30 fixed labels (different set of labels for each task). 

\paragraph{Quality Assurance and procedural step.}
Both tasks answer Yes/No questions about a transcript. \emph{Quality Assurance} asks whether the call satisfies a service-quality criterion (1--10 questions per conversation). \emph{Procedural Step} asks whether the agent carried out a specific step, such as verifying the caller's identity (2--20 questions per conversation).

\paragraph{Intent classification.}
We evaluate whether models can correctly classify the intent of the customer utterance. Unlike call outcome and purpose of call, where 30 fixed labels are always provided, the candidate labels in intent classification are dynamically provided in the input, ranging from 4 to 21 options. 

\paragraph{RAG document selection.}
Given a conversation and three {knowledge-base} articles, this task evaluates whether models can select the most relevant article that can answer the customer's question, or $-1$ if none does.

\subsection{Judging Tasks}
\label{sec:judging}
\paragraph{Agent workflow judging.}
This dataset contains customer-agent conversations where the LLM agent is required to resolve a support issue by following a skill workflow \cite{laskar2026evaluating}. The judge receives the instructions, policies, conversation turns, and tool call logs. It {returns a {five-way task-completion verdict (success, failure, partial, appropriately transferred, or appropriately detecting out of scope)}, a yes/no verdict for hallucinated actions, and five 1--5 rubric ratings} for the following criteria: workflow adherence, tool-use correctness, guardrail compliance, communication quality, and resolution efficiency. An overall pass verdict is then derived by a fixed rule rather than asked directly: a conversation passes if the task completion is a success or an appropriate transfer or out-of-scope verdict, workflow adherence is at least 4, no hallucinated action is found. 

\paragraph{Model-generated response judging.}
Judges evaluate summaries and action items generated by two models, Qwen-3.5-9B \cite{qwen3.5} and Qwen-3.8-27B \cite{qwen3.8}. We evaluate the responses using both pointwise and pairwise metrics. For {pointwise}, each response is rated from 1 to 5 for faithfulness, completeness, and instruction following. For pairwise, the judge chooses the better response, or a tie. To measure position bias, each pair is shown in both orders.

\begin{table*}[t]
\centering\setlength{\tabcolsep}{4pt}\renewcommand{\arraystretch}{0.94}
\begin{adjustbox}{max width=\textwidth}\begin{tabular}{llccccccc}
\toprule
Model & Size & Call Outcome & Purpose of Call & {Quality Assurance} & Procedural Step & Intent & RAG Doc. & Avg. \\
\midrule
\rowcolor{rowDecision} Gemini-Decision-1 & API & 51.5 & 76.3 & 85.4 & 86.4 & \textbf{84.0} & {\textbf{73.4}} & {\textbf{76.2}} \\
\rowcolor{rowDecision} OpenJev & 27B & 50.8 & 78.6 & 87.3 & 86.6 & 79.9 & {72.4} & {75.9} \\
\rowcolor{rowDecision} GPT-6-Luna-Decisions & API & \textbf{54.6} & \textbf{80.3} & 87.0 & 85.4 & 77.6 & {70.5} & 75.9 \\
\rowcolor{rowDecision} JEV & API & 52.7 & 79.8 & 86.6 & \textbf{87.1} & 78.7 & {68.2} & 75.5 \\
\rowcolor{rowDecision} Clef & 27B & 52.1 & 78.6 & \textbf{87.5} & 84.8 & 76.2 & {62.3} & 73.6 \\
\rowcolor{rowDecision} OpenJev-Flash & 9B & 44.8 & 76.9 & 82.0 & 84.0 & 73.4 & {65.6} & 71.1 \\
\rowcolor{rowDecision} Clef-Flash & 9B & 45.8 & 74.0 & 81.1 & 83.6 & 71.3 & {65.8} & 70.2 \\
\rowcolor{rowDecision} lev & 4B & 42.2 & 66.5 & 77.0 & 83.0 & 72.2 & {61.4} & {67.1} \\
\rowcolor{rowDecision} Kev & 4B & 27.1 & 45.7 & 81.5 & 82.7 & 69.3 & {61.6} & {61.3} \\
\midrule
\rowcolor{rowLLM} Qwen-3.8-27B & 27B & \textbf{55.2} & 76.9 & \textbf{88.1} & 87.2 & 85.4 & {\textbf{79.0}} & {\textbf{78.6}} \\
\rowcolor{rowLLM} Gemini-3.8-Flash & API & 54.4 & 76.9 & 84.2 & 86.7 & \textbf{86.1} & {76.3} & 77.4 \\
\rowcolor{rowLLM} GPT-6-Luna & API & 51.1 & 76.9 & 86.9 & 87.0 & 81.8 & {69.9} & {75.6} \\
\rowcolor{rowLLM} Gemma-4-26B-A4B & 26B & 53.6 & 71.1 & 87.2 & 86.2 & 79.7 & {75.5} & {75.5} \\
\rowcolor{rowLLM} Gemini-3.5-Flash-Lite & API & 50.6 & 79.2 & 86.3 & \textbf{87.5} & 80.1 & {69.5} & {75.5} \\
\rowcolor{rowLLM} GPT-5.6-Luna & API & 52.7 & 79.2 & 85.8 & 86.7 & 80.7 & {63.3} & 74.7 \\
\rowcolor{rowLLM} Claude-Haiku-5.5 & API & 53.8 & \textbf{82.7} & 82.7 & 83.5 & 78.9 & {61.6} & {73.9} \\
\rowcolor{rowLLM} Qwen-3.6-35B-A3B & 35B & 52.3 & 74.6 & 87.5 & 84.6 & 75.5 & {67.4} & {73.6} \\
\rowcolor{rowLLM} Gemma-4-E4B & 8B & 51.1 & 76.3 & 83.8 & 85.9 & 78.5 & {62.1} & {73.0} \\
\rowcolor{rowLLM} Qwen-3.5-9B & 9B & 47.5 & 74.6 & 85.7 & 85.5 & 71.5 & 61.0 & 71.0 \\
\rowcolor{rowLLM} Gemma-4-E2B & 5B & 38.0 & 60.1 & 78.2 & 79.7 & 69.9 & {47.0} & 62.2 \\
\rowcolor{rowLLM} Gemma-3-4B & 4B & 34.5 & 65.3 & 73.0 & 74.7 & 65.9 & {38.6} & 58.7 \\
\rowcolor{rowLLM} Qwen-3.5-2B & 2B & 31.9 & 52.0 & 76.2 & 76.7 & 62.9 & {39.8} & {56.6} \\
\bottomrule
\end{tabular}
\end{adjustbox}
\caption{{Accuracy (\%) on the six classification tasks; Avg.\ is the unweighted mean.}} 
\label{tab:classification}
\end{table*}

\begin{table*}[t]
\centering\setlength{\tabcolsep}{5.5pt}\renewcommand{\arraystretch}{0.94}
\begin{adjustbox}{max width=\textwidth}\begin{tabular}{llccccccc}
\toprule
 & & \multicolumn{3}{c}{Reference: GPT-6.1-Sol} & \multicolumn{3}{c}{Reference: Claude-Sonnet-5.5} & Overall \\
\cmidrule(lr){3-5}\cmidrule(lr){6-8}
 & & \multicolumn{2}{c}{Generated response} & Agent Workflow & \multicolumn{2}{c}{Generated response} & Agent Workflow & \\
Judge & Size & Pointwise & Pairwise & Avg. & Pointwise & Pairwise & Avg. & Avg. \\
\midrule
\rowcolor{rowDecision} GPT-6-Luna-Decisions & API & \textbf{0.55} & \textbf{0.39} & {0.61} & \textbf{0.60} & 0.33 & {0.66} & {\textbf{0.52}} \\
\rowcolor{rowDecision} JEV & API & 0.47 & 0.38 & {0.66} & 0.51 & \textbf{0.36} & {0.66} & {0.51} \\
\rowcolor{rowDecision} Clef & 27B & 0.49 & 0.19 & {\textbf{0.68}} & 0.54 & 0.26 & {\textbf{0.70}} & {0.48} \\
\rowcolor{rowDecision} OpenJev & 27B & 0.47 & 0.25 & {0.57} & 0.44 & {0.33} & {0.63} & {0.45} \\
\rowcolor{rowDecision} Gemini-Decision-1 & API & 0.34 & 0.37 & {0.58} & 0.31 & {0.32} & {0.57} & {0.42} \\
\rowcolor{rowDecision} Clef-Flash & 9B & 0.25 & $-$0.07 & {0.35} & 0.28 & $-$0.02 & {0.37} & {0.19} \\
\rowcolor{rowDecision} OpenJev-Flash & 9B & 0.04 & $-$0.02 & {0.40} & 0.07 & 0.05 & {0.45} & {0.17} \\
\rowcolor{rowDecision} Kev & 4B & 0.02 & $-$0.01 & {0.28} & 0.02 & 0.03 & {0.33} & {0.11} \\
\rowcolor{rowDecision} lev & 4B & $-$0.01 & 0.04 & {0.13} & 0.03 & 0.00 & {0.16} & {0.06} \\
\midrule
\rowcolor{rowLLM} Gemini-3.8-Flash & API & 0.57 & \textbf{0.64} & {0.73} & 0.47 & \textbf{0.48} & {0.70} & {\textbf{0.60}} \\
\rowcolor{rowLLM} GPT-5.6-Luna & API & 0.63 & 0.47 & {\textbf{0.78}} & 0.58 & 0.37 & {\textbf{0.74}} & {0.59} \\
\rowcolor{rowLLM} GPT-6-Luna & API & \textbf{0.66} & 0.56 & {0.58} & \textbf{0.59} & 0.44 & {0.65} & {0.58} \\
\rowcolor{rowLLM} Qwen-3.8-27B & 27B & 0.45 & 0.42 & {0.69} & 0.45 & 0.39 & {0.72} & {0.52} \\
\rowcolor{rowLLM} Claude-Haiku-5.5 & API & 0.44 & 0.18 & {0.62} & 0.51 & 0.16 & {0.66} & {0.43} \\
\rowcolor{rowLLM} Gemini-3.5-Flash-Lite & API & 0.28 & 0.34 & {0.57} & 0.28 & 0.29 & {0.59} & {0.39} \\
\rowcolor{rowLLM} Qwen-3.6-35B-A3B & 35B & 0.29 & 0.23 & {0.58} & 0.28 & 0.21 & {0.62} & {0.37} \\
\rowcolor{rowLLM} Gemma-4-26B-A4B & 26B & 0.20 & 0.35 & {0.55} & 0.19 & 0.31 & {0.56} & {0.36} \\
\rowcolor{rowLLM} Qwen-3.5-9B & 9B & 0.07 & 0.05 & {0.53} & 0.11 & 0.05 & {0.55} & {0.23} \\
\rowcolor{rowLLM} Gemma-4-E4B & 8B & 0.21 & 0.04 & {0.35} & 0.23 & 0.12 & {0.38} & {0.22} \\
\rowcolor{rowLLM} Gemma-4-E2B & 5B & 0.18 & $-$0.06 & {0.40} & 0.22 & $-$0.05 & {0.43} & {0.19} \\
\rowcolor{rowLLM} Gemma-3-4B & 4B & $-$0.05 & $-$0.05 & {0.30} & $-$0.02 & $-$0.04 & {0.32} & {0.08} \\
\rowcolor{rowLLM} Qwen-3.5-2B & 2B & $-$0.07 & $-$0.01 & {0.12} & $-$0.06 & $-$0.02 & {0.15} & {0.02} \\
\bottomrule
\end{tabular}
\end{adjustbox}
\caption{ Agreement with two reference judges.
\emph{Pointwise}: mean Spearman's $\rho$ across faithfulness, completeness, and instruction-following ratings of generated responses;
\emph{Pairwise}: Cohen's $\kappa$ on preference verdicts;
\emph{Agent Workflow Avg.}: the average of seven scores: Cohen's $\kappa$ for pass and task-completion verdicts and Spearman's $\rho$ for each of the five rubric ratings.
{\emph{Overall Avg.}: the average of the six displayed scores: pointwise, pairwise, and agent workflow for each of the two reference judges. 
{Appendix~\ref{app:judging} provides the individual scores, raw agreement, and rating differences.}}}
\label{tab:judging}
\end{table*}

\section{Experimental Setup}
\label{sec:setup}
\paragraph{Models.}
We evaluate {nine} decision models: {the API-based closed} JEV~1.13~\citep{typesafe2026}, {Gemini-Decision-1~\citep{google2026decisions} and GPT-6-Luna-Decisions~\citep{openai2026decisions}}; the open-weight Clef~\citep{clef2026} and OpenJev~\citep{openjev2026} (27B) and their 9B Flash variants; lev and Kev~\citep{palmer2026kev4b} (4B). The generative baselines comprise {eight} open LLMs covering Qwen~\citep{qwen3.5,qwen3.6-35b-a3b,qwen3.8} and Gemma~\citep{gemmateam2025gemma3,gemmateam2026gemma4} series models and {five} cost-effective API models: {GPT-5.6-Luna}~\citep{openai2026gpt56luna}, {GPT-6-Luna}~\citep{openaiprices2026}, {Gemini-3.8-Flash}~\citep{google2026gemini38flash}, {Gemini-3.5-Flash-Lite}~\citep{google2026gemini35flashlite}, and {Claude-Haiku-5.5}~\citep{anthropic2026haiku55}. {We use two reference judges from two model families: GPT-6.1-Sol~\citep{openai2026gpt61sol} (primary) and Claude-Sonnet-5.5~\citep{anthropic2026sonnet55}.} 

\paragraph{Protocol.}
An evaluation instance is a triplet $(x,q,\mathcal{Y})$: the task context $x$, a question $q$, and the allowed answers $\mathcal{Y}$. All models are evaluated zero-shot, without any fine-tuning or few-shot demonstrations. Descriptions of the candidate labels are also provided where applicable (i.e., call outcome, purpose of call, intent classification). Decision models receive $\mathcal{Y}$ as typed options and return an answer with option probabilities, while generative LLMs return the answer as structured output. {Appendix~\ref{app:prompts} shows prompt templates for all tasks.} 

\paragraph{Metrics.}
For classification, we report exact match accuracy by comparing the model-selected label with the gold label. For judging, we follow prior work that evaluates 
cheap smaller judges by comparing against a stronger one~\citep{laskar2025judging}. {Here, we use GPT-6.1-Sol and Claude-Sonnet-5.5 as reference judges and measure correlation between our evaluated models and reference judges in terms 
Cohen's $\kappa$ for categorical verdicts and Spearman's $\rho$ for 1--5 ratings.} 

\section{Results}
\label{sec:results}
\subsection{Classification Tasks}
\label{sec:results-classification}

\paragraph{Overall ranking.}
Table~\ref{tab:classification} shows that Qwen-3.8-27B achieves the highest average accuracy across the six classification tasks (78.6), followed by Gemini-3.8-Flash (77.4) and Gemini-Decision-1 (76.2). Beyond the top two models, performance is closely clustered: the nine models ranked third to eleventh differ by only 2.3 percentage points. Thus, although the strongest generative models lead overall, the performance of several decision models is also comparable to the generative LLMs.

\paragraph{Decision models.}
{Gemini-Decision-1 is the strongest decision model (76.2), followed by OpenJev and GPT-6-Luna-Decisions (both 75.9), JEV (75.5), and Clef (73.6). These leading decision models also outperform many generative LLMs, indicating that free-form generation is not necessary for strong performance on these bounded classification tasks. Among open-weight decision models, larger variants generally achieve higher accuracy. However, performance differences remain pronounced on Intent classification and RAG document selection. Averaged over these two tasks, OpenJev (76.1), JEV (73.5), and Clef (69.3) trail Qwen-3.8-27B (82.2) and Gemini-3.8-Flash (81.2). Gemini-Decision-1 achieves 78.7, narrowing this gap and showing that the weakness is not shared equally across decision models.}

\paragraph{Task difficulty and scale.}
Call Outcome is the hardest task for all models (best: 55.2), whereas the strongest models reach 86--88\% in  Quality Assurance and Procedural Step. Accuracy increases with size within each family: moving from 9B to 27B adds {4.8} points for OpenJev and {3.3} for Clef. 
\subsection{Judging Tasks}
\label{sec:results-judging}
\paragraph{Overall ranking.}
{Gemini-3.8-Flash has the highest overall agreement (0.60), followed by GPT-5.6-Luna (0.59) and GPT-6-Luna (0.58). Meanwhile, Qwen-3.8-27B, which is the strongest classifier, reaches only 0.52.}
%

\paragraph{Decision models as judges.}
{GPT-6-Luna-Decisions leads among decision models (overall 0.52), followed by JEV (0.51) and Clef (0.48); Gemini-Decision-1 reaches only 0.42 despite leading decision-model across classification tasks. 
For the four strongest decision models, generated-response pointwise correlations range from 0.44 to 0.60, but pairwise $\kappa$ ranges from 0.19 to 0.39.}

\paragraph{Robustness to the reference judge.}
The judge rankings are nearly unchanged across references (Spearman $\rho=0.99$). Gemini-3.8-Flash leads under GPT-6.1-Sol (0.65), followed by GPT-5.6-Luna (0.63) and GPT-6-Luna (0.60). Under Claude-Sonnet-5.5, GPT-5.6-Luna and GPT-6-Luna lead (both 0.56), followed by Gemini-3.8-Flash (0.55). This pattern suggests that same-family preference is not a major driver of the overall rankings: the GPT judges remain competitive under the Claude reference, while the Gemini judge leads under the GPT reference. 


\paragraph{Effect of scale.}
{Judging is more sensitive to scale than classification. Against GPT-6.1-Sol, the 9B decision models retain less than half the average agreement of their 27B counterparts: 0.18 vs.\ 0.45 for Clef and 0.14 vs.\ 0.43 for OpenJev. The 4B models perform even worse (Kev 0.10, lev 0.05).}

\begin{figure*}[!t]
\centering\includegraphics[width=\textwidth]{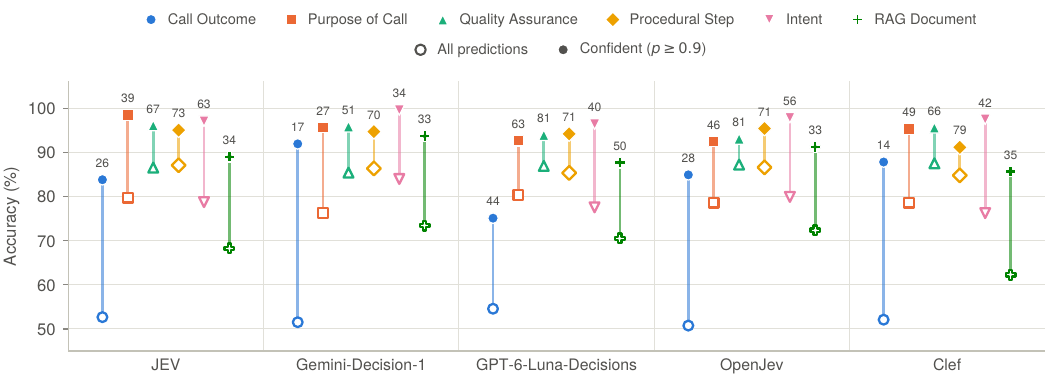}
\caption{{Accuracy of the five {strongest classification} decision models on all predictions (empty markers) and on confident predictions with selected-option probability $p\geq0.9$ (filled markers), for each classification task. The number above each pair is the percentage of predictions that are confident.}}
\label{fig:confidence}
\end{figure*}

\begin{table}[t]
\centering\setlength{\tabcolsep}{3.5pt}\renewcommand{\arraystretch}{0.92}
\begin{adjustbox}{max width=0.95\columnwidth}\begin{tabular}{lcccc}
\toprule
 & \multicolumn{2}{c}{Intent} & \multicolumn{2}{c}{RAG Document} \\
\cmidrule(lr){2-3}\cmidrule(lr){4-5}
Model & Answ. & Abst.-R & Answ. & Abst.-R \\
\midrule
\rowcolor{rowDecision} Gemini-Decision-1 & 83.3 & \textbf{86.8} & {78.7} & {\textbf{64.4}} \\
\rowcolor{rowDecision} OpenJev & \textbf{92.2} & 27.1 & {\textbf{86.2}} & {48.9} \\
\rowcolor{rowDecision} GPT-6-Luna-Decisions & 89.8 & 25.0 & {76.7} & {60.0} \\
\rowcolor{rowDecision} JEV & 92.1 & 21.5 & {82.0} & {45.0} \\
\rowcolor{rowDecision} Clef & 89.8 & 18.1 & {83.9} & {25.6} \\
\rowcolor{rowDecision} OpenJev-Flash & 88.2 & 9.7 & {84.3} & {33.9} \\
\rowcolor{rowDecision} Clef-Flash & 86.4 & 6.3 & {77.4} & {46.1} \\
\rowcolor{rowDecision} lev & 85.0 & 17.4 & {78.7} & {32.2} \\
\rowcolor{rowDecision} Kev & 80.7 & 20.1 & {77.0} & {35.6} \\
\midrule
\rowcolor{rowLLM} Qwen-3.8-27B & 86.4 & 81.3 & {80.3} & {\textbf{76.7}} \\
\rowcolor{rowLLM} Gemini-3.8-Flash & 86.4 & \textbf{84.7} & {77.7} & {73.9} \\
\rowcolor{rowLLM} GPT-6-Luna & 87.5 & 56.9 & {83.0} & {47.8} \\
\rowcolor{rowLLM} Gemma-4-26B-A4B & \textbf{91.4} & 29.2 & {83.0} & {62.8} \\
\rowcolor{rowLLM} Gemini-3.5-Flash-Lite & 90.3 & 36.1 & {79.0} & {53.3} \\
\rowcolor{rowLLM} GPT-5.6-Luna & 89.8 & 41.7 & {\textbf{84.3}} & {27.8} \\
\rowcolor{rowLLM} Claude-Haiku-5.5 & 90.5 & 29.2 & {77.7} & {34.4} \\
\rowcolor{rowLLM} Qwen-3.6-35B-A3B & 87.7 & 22.9 & {75.1} & {54.4} \\
\rowcolor{rowLLM} Gemma-4-E4B & 88.7 & 34.7 & {81.0} & {30.0} \\
\rowcolor{rowLLM} Qwen-3.5-9B & 85.6 & 11.1 & {74.4} & {38.3} \\
\rowcolor{rowLLM} Gemma-4-E2B & 85.3 & 4.2 & {74.8} & 0.0 \\
\rowcolor{rowLLM} Gemma-3-4B & 80.1 & 4.9 & {59.3} & {3.3} \\
\rowcolor{rowLLM} Qwen-3.5-2B & 77.3 & 0.7 & {63.0} & {0.6} \\
\bottomrule
\end{tabular}
\end{adjustbox}
\caption{{Answering versus abstaining (\%). \emph{Answ.}: accuracy on inputs that have an answer (618 Intent utterances with a configured intent; {305} RAG conversations with a relevant article). \emph{Abst.-R}: recall of the abstention option (\textit{no intent} for Intent, ``none relevant'' for RAG).}}
\label{tab:abstention-main}
\end{table}
\subsection{Analysis}
\label{sec:analysis}
\paragraph{Classification versus judging.}
Across the {22} models, rankings of models between classification and judging are strongly but not perfectly correlated (Spearman $\rho={0.85}$). Notably, GPT-5.6-Luna rises from {tenth} as a classifier to {second} as a judge, whereas Gemini-Decision-1 drops from third to tenth. Strong performance in classification therefore does not imply similar performance in judging. 

\paragraph {Decision Models Rarely Abstain.}
\label{sec:abstention}
The largest source of error for {most} decision models is a reluctance to choose the no-answer option ({see Table~\ref{tab:abstention-main}}). On Intent, OpenJev (92.2\%) and JEV (92.1\%) are the most accurate of all models on requests that match a configured intent, above Gemini-3.8-Flash and Qwen-3.8-27B (both 86.4\%). However, {eight of the nine decision models in Table~\ref{tab:classification}} could correctly classify only 6--27\% of \textit{no intent} requests, compared with 85\% for Gemini-3.8-Flash and 81\% for Qwen-3.8-27B. 
Gemini-Decision-1 is the exception: it correctly classifies 86.8\% of \textit{no intent} requests. 
RAG document selection shows the same pattern. When a relevant article exists, {JEV, OpenJev, and Clef} correctly select an acceptable one in {82--86\% of cases, above Qwen-3.8-27B (80.3\%) and Gemini-3.8-Flash (77.7\%)}. When none is relevant, they answer ``none'' in only {26--49\% of cases, compared with 77\% and 74\% for the respective LLMs}. 
Some LLMs share this failure (e.g., {28\%} for GPT-5.6-Luna), and some smaller LLMs almost never abstain (e.g., Gemma-3-4B, Qwen-3.5-2B). 

\paragraph{Confidence as a Routing Signal.}
\label{sec:confidence}
Decision models return a probability $p$ for the selected option. Across six classification tasks, confidence predicts correctness with AUROC \cite{hanley1982meaning} 0.69--0.92 for the seven decision models of at least 9B or accessed through APIs. Figure~\ref{fig:confidence} shows accuracy when confidence is at $p\geq0.9$ for the five best decision models. On Intent, JEV achieves 97.1\% accuracy at $p\geq0.9$, compared with 78.7\% overall. At $p\geq0.9$, Gemini-Decision-1 reaches 99.6\%. On Quality Assurance and Procedural Step, JEV, OpenJev, and Clef achieve 91--96\% accuracy at $p\geq0.9$. These results support accepting confident predictions and routing the remainder to a stronger model.
However, confidence has {two} limitations. Confidence coverage with $p\geq0.9$ varies by task: on Call Outcome, it is only 14--28\% for JEV, OpenJev, Clef, and Gemini-Decision-1; GPT-6-Luna-Decisions reaches 44\% coverage but only 75\% accuracy. Confident errors also persist when abstention is required: 12 of JEV's 14 confident Intent errors assign an intent to \textit{no intent} requests. 


\paragraph{Position and order sensitivity.}
We test five leading pairwise judges for position bias, defined as selecting the same displayed position after reversing response order (Table~\ref{tab:position}). JEV selects the second position in both orders for 24.6\% of pairs, versus 7.2\% for GPT-6-Luna and 1.5\% for Gemini-3.8-Flash. Conversely, GPT-6-Luna-Decisions and Gemini-Decision-1 favor the first position for 45.6\% and 26.1\% of pairs, consistently selecting the same response across orders in only 48.9\% and 56.1\%. 
We also find that high confidence ($p\geq0.9$ ) weakly predicts judging correctness: pairwise-preference AUROC is only 0.57--0.66 for JEV, OpenJev, and Clef.

\begin{table}[t]
\centering\setlength{\tabcolsep}{4pt}
\begin{adjustbox}{max width=\columnwidth}\begin{tabular}{lccc}
\toprule
Judge & Same resp. & Always 1st & Always 2nd \\
\midrule
\rowcolor{rowDecision} JEV & 67.9 & 1.8 & 24.6 \\
\rowcolor{rowLLM} Gemini-3.8-Flash & 83.5 & 7.2 & 1.5 \\
\rowcolor{rowLLM} GPT-6-Luna & 72.6 & 12.2 & 7.2 \\
\rowcolor{rowDecision} {Gemini-Decision-1} & {56.1} & {26.1} & {0.9} \\
\rowcolor{rowDecision} {GPT-6-Luna-Decisions} & {48.9} & {45.6} & {0.3} \\
\bottomrule
\end{tabular}
\end{adjustbox}
\caption{{Position bias in pairwise judging of model-generated responses.  \emph{Same resp.}: the judge prefers the same response in both orders (higher is ideal). \emph{Always 1st/2nd}: the judge picks the same displayed position in both orders (lower is ideal).}}
\label{tab:position}
\end{table}

\section{Cost and Deployment Guidance}
\label{sec:cost}
\begin{table}[t]
\centering\setlength{\tabcolsep}{5pt}\renewcommand{\arraystretch}{0.94}
\begin{adjustbox}{max width=\columnwidth}\begin{tabular}{llrrr}
\toprule
Model & Serving & p50 (s) & \$/1k & Acc. \\
\midrule
\rowcolor{rowDecision} lev & local & 0.40 & -- & {67.1} \\
\rowcolor{rowDecision} OpenJev-Flash & local & 0.40 & -- & 71.1 \\
\rowcolor{rowDecision} Clef-Flash & local & 0.37 & -- & 70.2 \\
\rowcolor{rowDecision} Clef & local & 1.21 & -- & 73.6 \\
\rowcolor{rowDecision} OpenJev & local & 1.22 & -- & {75.9} \\
\rowcolor{rowLLM} Qwen-3.8-27B & local & 1.58 & -- & {78.6} \\
\midrule
\rowcolor{rowDecision} {JEV} & {API} & {--} & {0.09} & {75.5} \\
\rowcolor{rowDecision} Gemini-Decision-1 & API & -- & 0.20 & {76.2} \\
\rowcolor{rowDecision} GPT-6-Luna-Decisions & API & -- & 0.23 & 75.9 \\
\rowcolor{rowLLM} GPT-6-Luna & API & -- & 0.29 & {75.6} \\
\rowcolor{rowLLM} Claude-Haiku-5.5 & API & -- & 0.34 & {73.9} \\
\rowcolor{rowLLM} GPT-5.6-Luna & API & -- & 0.58 & 74.7 \\
\rowcolor{rowLLM} Gemini-3.5-Flash-Lite & API & -- & 0.68 & {75.5} \\
\rowcolor{rowLLM} Gemini-3.8-Flash & API & -- & 1.73 & 77.4 \\
\bottomrule
\end{tabular}
\end{adjustbox}
\caption{{{Estimated cost per 1,000 requests and median latency per request.} Local models use one NVIDIA A100 GPU (p50: median latency), and therefore no usage cost is reported; API costs use recorded usage and {list prices of November 2026; the API-based closed decision models (JEV, Gemini-Decision-1, GPT-6-Luna-Decisions) do not bill output tokens}. Acc.\ is the average from Table~\ref{tab:classification}.}}
\label{tab:cost}
\end{table}

\paragraph{Cost.}
{JEV is the cheapest measured judge at} \$0.10 per 1,000 judgments, versus \$0.22 for Gemini-Decision-1, \$0.25 for GPT-6-Luna-Decisions, \$0.32 for GPT-6-Luna, and \$2.00 for Gemini-3.8-Flash. However, its $\kappa$ is lower than those of the two generative models (0.38 vs.\ 0.56 and 0.64). For classification, the three decision APIs cost \$0.09--0.23 per 1,000 requests, with similar or higher accuracy than GPT-6-Luna, the cheapest generative API at \$0.29. Among open-weight models (benchmarked on one A100 GPU), 9B decision models are the fastest (median 0.4\,s), but 27B models are 3--4x slower. 

\paragraph{Recommendations.}
{Before deployment, (1)~evaluate classification and judging separately; (2)~measure no-answer recall; (3)~calibrate confidence thresholds for classification routing; (4)~validate pointwise and workflow verdicts; and (5)~evaluate pairwise comparisons in both orders.}
\section{Conclusion}
We evaluated {nine} decision models and {thirteen} generative LLMs on eight real-world {contact center} tasks. Many decision models can match frontier LLM performance on call classification, are competitive judges of multi-turn agent workflows, and expose confidence scores that enable selective routing; API-based closed decision models also have the lowest per-request cost among the API models we evaluated. However, {most} rarely abstain on unsupported intents and irrelevant documents, judge pairwise preferences poorly {and with position bias}, and lose most of their judging quality at smaller scale. Model selection should therefore rely on abstention tests, judge agreement, confidence audits, and measured cost alongside overall accuracy. Future work will further study calibrated abstention thresholds and evaluate across more tasks.
\label{main-end}

\section*{Limitations}
\noindent 
Our study is offline and measures no live customer outcomes. The four call-level tasks use stored operational labels whose annotation process is undocumented, and the Intent utterances are synthetic. 
The confidence analysis is retrospective and uses a fixed threshold. 
Generalization to other settings, organizations, languages, and future model versions remains untested.

\section*{Ethical Considerations}
Incorrect {contact center} decisions can misroute customers, overlook procedure failures, or pass irrelevant evidence to an answer generator. We therefore report abstention recall and confident errors, because aggregate accuracy can hide these risks, and we recommend human review before consequential use. The data come from private {contact center} customer interactions that are PII-redacted. Licensing requirements are followed accordingly for the evaluated models.

\bibliography{references}
\appendix

\section{Additional Classification Results}
\label{app:classification}
\paragraph{Macro-F1.}
Table~\ref{tab:f1} reports macro-F1 on the four call-level tasks for all {22} models that completed them. The ranking is similar to that by accuracy: {GPT-6-Luna-Decisions has the best average macro-F1 of all models (75.5), and JEV (74.1) is on par with the best generative LLMs (74.3)}.

\begin{table*}[!tbp]
\centering\small\setlength{\tabcolsep}{5pt}
\begin{adjustbox}{max width=\textwidth}\begin{tabular}{llcccccc}
\toprule
Model & Size & Call Outcome & Purpose of Call & {Quality Assurance} & Procedural Step & Avg. F1 & Avg. Acc. \\
\midrule
\rowcolor{rowDecision} {GPT-6-Luna-Decisions} & {API} & {\textbf{51.9}} & {\textbf{81.1}} & {86.8} & {82.2} & {\textbf{75.5}} & {\textbf{76.8}} \\
\rowcolor{rowDecision} JEV & API & 49.9 & 76.8 & 86.5 & \textbf{83.1} & 74.1 & 76.5 \\
\rowcolor{rowDecision} OpenJev & 27B & 47.5 & 79.0 & 87.0 & 82.3 & 73.9 & 75.8 \\
\rowcolor{rowDecision} {Gemini-Decision-1} & {API} & {48.6} & {78.5} & {85.4} & {80.2} & {73.2} & {74.9} \\
\rowcolor{rowDecision} Clef & 27B & 47.2 & 79.1 & \textbf{87.5} & 76.3 & 72.5 & 75.8 \\
\rowcolor{rowDecision} OpenJev-Flash & 9B & 41.2 & 79.2 & 81.2 & 80.4 & 70.5 & 72.0 \\
\rowcolor{rowDecision} Clef-Flash & 9B & 43.4 & 75.2 & 81.0 & 74.6 & 68.6 & 71.1 \\
\rowcolor{rowDecision} lev & 4B & 40.2 & 65.9 & 75.1 & 77.9 & 64.8 & 67.2 \\
\rowcolor{rowDecision} Kev & 4B & 27.4 & 44.7 & 81.1 & 77.4 & 57.7 & 59.2 \\
\midrule
\rowcolor{rowLLM} GPT-5.6-Luna & API & 49.4 & 78.6 & 85.6 & \textbf{83.7} & \textbf{74.3} & 76.1 \\
\rowcolor{rowLLM} Qwen-3.8-27B & 27B & \textbf{51.1} & 76.7 & \textbf{88.1} & 81.3 & \textbf{74.3} & \textbf{76.8} \\
\rowcolor{rowLLM} Gemini-3.5-Flash-Lite & API & 47.8 & 79.6 & 86.2 & 82.8 & 74.1 & 75.9 \\
\rowcolor{rowLLM} Claude-Haiku-5.5 & API & 51.0 & \textbf{82.0} & 82.7 & 80.2 & 74.0 & 75.7 \\
\rowcolor{rowLLM} Gemini-3.8-Flash & API & 50.5 & 78.3 & 84.2 & 81.4 & 73.6 & 75.5 \\
\rowcolor{rowLLM} GPT-6-Luna & API & 48.3 & 75.7 & 86.7 & 83.5 & 73.6 & 75.5 \\
\rowcolor{rowLLM} Gemma-4-E4B & 8B & 49.0 & 75.2 & 83.6 & 80.9 & 72.1 & 74.3 \\
\rowcolor{rowLLM} Qwen-3.6-35B-A3B & 35B & 49.2 & 75.2 & 87.4 & 76.3 & 72.0 & 74.7 \\
\rowcolor{rowLLM} Gemma-4-26B-A4B & 26B & 50.7 & 70.1 & 87.1 & 80.1 & 72.0 & 74.5 \\
\rowcolor{rowLLM} Qwen-3.5-9B & 9B & 43.9 & 74.2 & 85.6 & 79.7 & 70.8 & 73.3 \\
\rowcolor{rowLLM} Gemma-4-E2B & 5B & 36.3 & 58.5 & 78.2 & 65.2 & 59.6 & 64.0 \\
\rowcolor{rowLLM} Gemma-3-4B & 4B & 31.5 & 64.3 & 71.4 & 69.6 & 59.2 & 61.9 \\
\rowcolor{rowLLM} Qwen-3.5-2B & 2B & 28.4 & 53.3 & 75.0 & 60.5 & 54.3 & 59.2 \\
\bottomrule
\end{tabular}
\end{adjustbox}
\caption{Macro-F1 ($\times100$) on the four call-level tasks, with the corresponding average accuracy for comparison. Bold marks the best value per column within a block.}
\label{tab:f1}
\end{table*}

\section{Additional Judging Results}
\label{app:judging}

\paragraph{{Per-score agreement for both reference judges.}}
{Tables~\ref{tab:judging-sol} and~\ref{tab:judging-sonnet} give all seven agreement scores behind Table~\ref{tab:judging} for each reference judge. The two references agree with each other at 0.62, and the judge rankings under the two references correlate at Spearman $\rho={0.99}$.}

{Generated-response judging uses 957 test response pairs from 426 source-transcript groups: 1,914 candidate responses for pointwise ratings and 1,914 ordered pairwise judgments.}

\paragraph{Raw agreement and rating distances.}
Tables~\ref{tab:judge-generated} and~\ref{tab:judge-skills} 
demonstrate raw agreement, signed rating bias (candidate minus reference), mean absolute error (MAE), and the percentage of ratings within one point of the reference between each evaluated model and the reference judge GPT-6.1-Sol. 

\begin{table*}[!tbp]
\centering\small\setlength{\tabcolsep}{5.5pt}\renewcommand{\arraystretch}{0.94}
\begin{adjustbox}{max width=\textwidth}\begin{tabular}{llcccccccc}
\toprule
 & & \multicolumn{4}{c}{Generated responses} & \multicolumn{3}{c}{Agent workflow} & \\
\cmidrule(lr){3-6}\cmidrule(lr){7-9}
Judge & Size & Preference & Faithful. & Complete. & Instruction & Pass & Task comp. & Rubric & Avg. \\
 & & $\kappa$ & $\rho$ & $\rho$ & $\rho$ & $\kappa$ & $\kappa$ & $\rho$ & \\
\midrule
\rowcolor{rowDecision} GPT-6-Luna-Decisions & API & \textbf{0.39} & \textbf{0.68} & \textbf{0.61} & 0.37 & \textbf{0.54} & 0.50 & 0.64 & {\textbf{0.52}} \\
\rowcolor{rowDecision} JEV & API & 0.38 & 0.66 & 0.35 & 0.41 & 0.53 & \textbf{0.60} & 0.70 & {0.50} \\
\rowcolor{rowDecision} Clef & 27B & 0.19 & 0.46 & 0.56 & \textbf{0.46} & 0.49 & 0.58 & \textbf{0.74} & {0.45} \\
\rowcolor{rowDecision} OpenJev & 27B & 0.25 & 0.62 & 0.43 & 0.36 & 0.43 & 0.58 & 0.60 & {0.43} \\
\rowcolor{rowDecision} Gemini-Decision-1 & API & 0.37 & 0.43 & 0.25 & 0.33 & 0.35 & 0.49 & 0.64 & {0.43} \\
\rowcolor{rowDecision} Clef-Flash & 9B & $-$0.07 & 0.30 & 0.26 & 0.21 & 0.10 & 0.36 & 0.40 & {0.18} \\
\rowcolor{rowDecision} OpenJev-Flash & 9B & $-$0.02 & 0.11 & 0.02 & $-$0.02 & 0.40 & 0.41 & 0.40 & {0.14} \\
\rowcolor{rowDecision} Kev & 4B & $-$0.01 & $-$0.07 & 0.05 & 0.07 & 0.17 & 0.31 & 0.29 & {0.10} \\
\rowcolor{rowDecision} lev & 4B & 0.04 & $-$0.02 & $-$0.05 & 0.03 & 0.15 & 0.42 & 0.07 & {0.05} \\
\midrule
\rowcolor{rowLLM} Gemini-3.8-Flash & API & \textbf{0.64} & 0.67 & 0.52 & 0.53 & 0.44 & 0.59 & 0.81 & {\textbf{0.65}} \\
\rowcolor{rowLLM} GPT-5.6-Luna & API & 0.47 & 0.73 & 0.64 & 0.52 & \textbf{0.65} & \textbf{0.69} & \textbf{0.82} & {0.63} \\
\rowcolor{rowLLM} GPT-6-Luna & API & 0.56 & \textbf{0.74} & \textbf{0.67} & \textbf{0.55} & 0.54 & 0.50 & 0.61 & {0.60} \\
\rowcolor{rowLLM} Qwen-3.8-27B & 27B & 0.42 & 0.51 & 0.45 & 0.39 & 0.48 & 0.57 & 0.76 & {0.52} \\
\rowcolor{rowLLM} Claude-Haiku-5.5 & API & 0.18 & 0.55 & 0.54 & 0.22 & 0.46 & 0.36 & 0.70 & {0.41} \\
\rowcolor{rowLLM} Gemini-3.5-Flash-Lite & API & 0.34 & 0.34 & 0.25 & 0.24 & 0.37 & 0.42 & 0.64 & {0.40} \\
\rowcolor{rowLLM} Gemma-4-26B-A4B & 26B & 0.35 & 0.31 & 0.17 & 0.12 & 0.27 & 0.47 & 0.62 & {0.37} \\
\rowcolor{rowLLM} Qwen-3.6-35B-A3B & 35B & 0.23 & 0.37 & 0.26 & 0.24 & 0.49 & 0.49 & 0.62 & {0.36} \\
\rowcolor{rowLLM} Qwen-3.5-9B & 9B & 0.05 & 0.20 & 0.11 & $-$0.08 & 0.44 & 0.42 & 0.56 & {0.22} \\
\rowcolor{rowLLM} Gemma-4-E4B & 8B & 0.04 & 0.24 & 0.30 & 0.10 & 0.18 & 0.37 & 0.39 & {0.20} \\
\rowcolor{rowLLM} Gemma-4-E2B & 5B & $-$0.06 & 0.22 & 0.26 & 0.07 & 0.28 & 0.30 & 0.45 & {0.17} \\
\rowcolor{rowLLM} Gemma-3-4B & 4B & $-$0.05 & $-$0.10 & $-$0.10 & 0.05 & 0.22 & 0.14 & 0.35 & {0.07} \\
\rowcolor{rowLLM} Qwen-3.5-2B & 2B & $-$0.01 & $-$0.15 & $-$0.04 & $-$0.02 & 0.13 & 0.18 & 0.10 & {0.01} \\
\bottomrule
\end{tabular}
\end{adjustbox}
\caption{{Agreement with GPT-6.1-Sol as the reference judge: Cohen's $\kappa$ for categorical verdicts and Spearman's $\rho$ for 1--5 ratings. Rubric is the mean $\rho$ over five rubric dimensions and {Avg.\ is the mean of the pointwise, pairwise, and agent-workflow scores for this reference, using the same component definitions as Table~\ref{tab:judging}.} Sorted by the average within each group; bold marks the best value per column within each group.}}
\label{tab:judging-sol}
\end{table*}

\begin{table*}[!tbp]
\centering\small\setlength{\tabcolsep}{5.5pt}\renewcommand{\arraystretch}{0.94}
\begin{adjustbox}{max width=\textwidth}\begin{tabular}{llcccccccc}
\toprule
 & & \multicolumn{4}{c}{Generated responses} & \multicolumn{3}{c}{Agent workflow} & \\
\cmidrule(lr){3-6}\cmidrule(lr){7-9}
Judge & Size & Preference & Faithful. & Complete. & Instruction & Pass & Task comp. & Rubric & Avg. \\
 & & $\kappa$ & $\rho$ & $\rho$ & $\rho$ & $\kappa$ & $\kappa$ & $\rho$ & \\
\midrule
\rowcolor{rowDecision} GPT-6-Luna-Decisions & API & 0.33 & \textbf{0.71} & 0.58 & 0.51 & \textbf{0.62} & 0.52 & 0.69 & {\textbf{0.53}} \\
\rowcolor{rowDecision} JEV & API & \textbf{0.36} & 0.68 & 0.42 & 0.44 & 0.59 & 0.51 & 0.70 & {0.51} \\
\rowcolor{rowDecision} Clef & 27B & 0.26 & 0.50 & \textbf{0.60} & \textbf{0.52} & 0.59 & \textbf{0.56} & \textbf{0.75} & {0.50} \\
\rowcolor{rowDecision} OpenJev & 27B & 0.33 & 0.61 & 0.41 & 0.32 & 0.49 & 0.50 & 0.69 & {0.47} \\
\rowcolor{rowDecision} Gemini-Decision-1 & API & 0.32 & 0.40 & 0.28 & 0.26 & 0.39 & 0.39 & 0.65 & {0.40} \\
\rowcolor{rowDecision} Clef-Flash & 9B & $-$0.02 & 0.33 & 0.31 & 0.20 & 0.16 & 0.30 & 0.43 & {0.21} \\
\rowcolor{rowDecision} OpenJev-Flash & 9B & 0.05 & 0.13 & 0.07 & 0.01 & 0.49 & 0.36 & 0.46 & {0.19} \\
\rowcolor{rowDecision} Kev & 4B & 0.03 & $-$0.06 & 0.09 & 0.03 & 0.23 & 0.34 & 0.35 & {0.13} \\
\rowcolor{rowDecision} lev & 4B & 0.00 & $-$0.02 & 0.07 & 0.06 & 0.20 & 0.34 & 0.11 & {0.07} \\
\midrule
\rowcolor{rowLLM} GPT-5.6-Luna & API & 0.37 & 0.68 & 0.63 & 0.42 & \textbf{0.62} & 0.55 & \textbf{0.80} & {\textbf{0.56}} \\
\rowcolor{rowLLM} GPT-6-Luna & API & 0.44 & \textbf{0.70} & \textbf{0.64} & \textbf{0.43} & \textbf{0.62} & 0.52 & 0.68 & {\textbf{0.56}} \\
\rowcolor{rowLLM} Gemini-3.8-Flash & API & \textbf{0.48} & 0.55 & 0.47 & 0.38 & 0.45 & 0.54 & 0.79 & {0.55} \\
\rowcolor{rowLLM} Qwen-3.8-27B & 27B & 0.39 & 0.51 & 0.54 & 0.31 & 0.53 & \textbf{0.58} & 0.78 & {0.52} \\
\rowcolor{rowLLM} Claude-Haiku-5.5 & API & 0.16 & 0.61 & 0.60 & 0.32 & 0.53 & 0.44 & 0.73 & {0.44} \\
\rowcolor{rowLLM} Gemini-3.5-Flash-Lite & API & 0.29 & 0.33 & 0.29 & 0.22 & 0.45 & 0.34 & 0.66 & {0.39} \\
\rowcolor{rowLLM} Qwen-3.6-35B-A3B & 35B & 0.21 & 0.37 & 0.29 & 0.19 & 0.58 & 0.43 & 0.66 & {0.37} \\
\rowcolor{rowLLM} Gemma-4-26B-A4B & 26B & 0.31 & 0.30 & 0.19 & 0.08 & 0.29 & 0.41 & 0.64 & {0.35} \\
\rowcolor{rowLLM} Gemma-4-E4B & 8B & 0.12 & 0.24 & 0.31 & 0.13 & 0.23 & 0.30 & 0.42 & {0.24} \\
\rowcolor{rowLLM} Qwen-3.5-9B & 9B & 0.05 & 0.19 & 0.17 & $-$0.02 & 0.49 & 0.39 & 0.59 & {0.24} \\
\rowcolor{rowLLM} Gemma-4-E2B & 5B & $-$0.05 & 0.26 & 0.26 & 0.13 & 0.34 & 0.22 & 0.49 & {0.20} \\
\rowcolor{rowLLM} Gemma-3-4B & 4B & $-$0.04 & $-$0.08 & $-$0.01 & 0.04 & 0.26 & 0.14 & 0.37 & {0.09} \\
\rowcolor{rowLLM} Qwen-3.5-2B & 2B & $-$0.02 & $-$0.14 & 0.02 & $-$0.05 & 0.17 & 0.18 & 0.13 & {0.02} \\
\bottomrule
\end{tabular}
\end{adjustbox}
\caption{{Agreement with Claude-Sonnet-5.5 as the reference judge, on the same instances and columns as Table~\ref{tab:judging-sol}. Sorted by the average within each group; bold marks the best value per column within each group.}}
\label{tab:judging-sonnet}
\end{table*}

\begin{table*}[!tbp]
\centering\small\setlength{\tabcolsep}{2.6pt}
\begin{adjustbox}{max width=\textwidth}\begin{tabular}{lcccccccccc}
\toprule
 & Pairwise & \multicolumn{3}{c}{Exact agreement (\%)} & \multicolumn{3}{c}{Signed bias} & & & \\
\cmidrule(lr){3-5}\cmidrule(lr){6-8}
Judge & agree. & Faith. & Compl. & Instr. & Faith. & Compl. & Instr. & MAE & W-1 & Valid \\
\midrule
\rowcolor{rowDecision} {GPT-6-Luna-Decisions} & {63.9} & {56.6} & {50.7} & {46.4} & {$-$0.07} & {$-$0.21} & {$-$0.58} & {0.60} & {89.5} & {100.0} \\
\rowcolor{rowDecision} JEV & 62.7 & 51.3 & 47.1 & 60.7 & +0.46 & +0.15 & $-$0.17 & 0.57 & 91.3 & 100.0 \\
\rowcolor{rowDecision} Clef & 54.1 & 46.8 & 55.9 & 49.7 & +0.16 & +0.08 & $-$0.49 & 0.58 & 91.9 & 100.0 \\
\rowcolor{rowDecision} OpenJev & 58.2 & 38.9 & 36.9 & 73.9 & +0.75 & +0.83 & +0.23 & 0.66 & 85.8 & 100.0 \\
\rowcolor{rowDecision} {Gemini-Decision-1} & {60.7} & {35.5} & {38.9} & {74.5} & {+0.92} & {+0.85} & {+0.30} & {0.72} & {81.0} & {100.0} \\
\rowcolor{rowDecision} Clef-Flash & 39.8 & 41.8 & 43.2 & 44.5 & +0.29 & +0.32 & $-$0.47 & 0.79 & 82.7 & 100.0 \\
\rowcolor{rowDecision} OpenJev-Flash & 42.7 & 33.6 & 34.8 & 72.7 & +1.05 & +0.99 & +0.36 & 0.82 & 76.5 & 100.0 \\
\rowcolor{rowDecision} Kev & 41.3 & 32.9 & 33.9 & 73.5 & +1.07 & +0.98 & +0.36 & 0.83 & 76.2 & 100.0 \\
\rowcolor{rowDecision} lev & 32.5 & 29.0 & 27.3 & 58.7 & +0.47 & +0.26 & $-$0.19 & 1.04 & 68.7 & 100.0 \\
\midrule
\rowcolor{rowLLM} GPT-5.6-Luna & 67.3 & 59.7 & 60.6 & 71.0 & +0.29 & +0.02 & $-$0.06 & 0.40 & 96.7 & 100.0 \\
\rowcolor{rowLLM} Gemini-3.8-Flash & 78.5 & 42.1 & 39.5 & 74.5 & +0.68 & +0.74 & +0.09 & 0.59 & 89.2 & 99.8 \\
\rowcolor{rowLLM} GPT-6-Luna & 73.7 & 66.0 & 62.6 & 74.0 & +0.06 & $-$0.05 & +0.00 & 0.35 & 97.4 & 100.0 \\
\rowcolor{rowLLM} Qwen-3.8-27B & 64.5 & 37.2 & 44.3 & 73.7 & +0.85 & +0.65 & +0.23 & 0.64 & 86.0 & 100.0 \\
\rowcolor{rowLLM} Claude-Haiku-5.5 & 48.2 & 40.9 & 33.8 & 16.0 & $-$0.43 & $-$0.65 & $-$1.19 & 0.93 & 78.9 & 100.0 \\
\rowcolor{rowLLM} Qwen-3.6-35B-A3B & 52.6 & 34.5 & 37.4 & 70.4 & +0.92 & +0.85 & +0.17 & 0.76 & 79.6 & 98.3 \\
\rowcolor{rowLLM} Gemini-3.5-Flash-Lite & 59.7 & 34.5 & 37.7 & 71.8 & +0.95 & +0.89 & +0.24 & 0.76 & 79.6 & 99.8 \\
\rowcolor{rowLLM} Gemma-4-26B-A4B & 60.0 & 33.7 & 34.8 & 73.8 & +1.02 & +0.98 & +0.37 & 0.79 & 77.4 & 100.0 \\
\rowcolor{rowLLM} Qwen-3.5-9B & 45.6 & 33.7 & 35.3 & 69.9 & +1.03 & +0.95 & +0.31 & 0.83 & 76.8 & 100.0 \\
\rowcolor{rowLLM} Gemma-4-E4B & 46.1 & 33.3 & 36.3 & 69.2 & +0.99 & +0.87 & +0.14 & 0.82 & 77.5 & 100.0 \\
\rowcolor{rowLLM} Gemma-4-E2B & 40.7 & 34.8 & 36.6 & 72.2 & +0.95 & +0.89 & +0.34 & 0.77 & 79.0 & 100.0 \\
\rowcolor{rowLLM} Gemma-3-4B & 41.0 & 30.6 & 33.0 & 74.0 & +1.00 & +0.91 & +0.39 & 0.83 & 76.3 & 100.0 \\
\rowcolor{rowLLM} Qwen-3.5-2B & 43.5 & 29.0 & 34.7 & 51.4 & +0.54 & +0.63 & +0.02 & 0.80 & 82.7 & 95.6 \\
\bottomrule
\end{tabular}
\end{adjustbox}
\caption{Model-generated response judging against GPT-6.1-Sol as reference: raw agreement, signed bias (candidate minus reference rating), MAE over the three rating dimensions, percentage of ratings within one point (W-1), and percentage of valid pointwise outputs. Pairwise and exact agreement are percentages.}
\label{tab:judge-generated}
\end{table*}

\begin{table*}[!tbp]
\centering\small\setlength{\tabcolsep}{3.4pt}
\begin{adjustbox}{max width=\textwidth}\begin{tabular}{lcccccccccc}
\toprule
 & \multicolumn{2}{c}{Agreement} & \multicolumn{5}{c}{Spearman $\rho$ per rubric dimension} & & & \\
\cmidrule(lr){2-3}\cmidrule(lr){4-8}
Judge & Pass & Task & Workflow & Tool use & Guardrail & Commun. & Efficiency & MAE & W-1 & Valid \\
\midrule
\rowcolor{rowDecision} JEV & 86.1 & 71.6 & 0.71 & 0.81 & 0.30 & 0.85 & 0.82 & 0.69 & 84.1 & 100.0 \\
\rowcolor{rowDecision} Clef & 90.0 & 69.0 & 0.74 & 0.77 & 0.52 & 0.83 & 0.84 & 0.59 & 88.7 & 100.0 \\
\rowcolor{rowDecision} {GPT-6-Luna-Decisions} & {90.2} & {64.5} & {0.66} & {0.61} & {0.44} & {0.77} & {0.73} & {0.74} & {83.7} & {100.0} \\
\rowcolor{rowDecision} OpenJev & 79.6 & 68.7 & 0.73 & 0.58 & 0.24 & 0.71 & 0.73 & 0.98 & 76.4 & 100.0 \\
\rowcolor{rowDecision} {Gemini-Decision-1} & {73.6} & {62.8} & {0.68} & {0.73} & {0.17} & {0.80} & {0.82} & {1.00} & {69.8} & {100.0} \\
\rowcolor{rowDecision} OpenJev-Flash & 83.5 & 57.7 & 0.47 & 0.47 & 0.15 & 0.47 & 0.44 & 1.13 & 71.1 & 100.0 \\
\rowcolor{rowDecision} Clef-Flash & 80.3 & 53.4 & 0.26 & 0.28 & 0.16 & 0.65 & 0.63 & 1.08 & 69.1 & 100.0 \\
\rowcolor{rowDecision} Kev & 80.4 & 52.1 & 0.13 & 0.35 & 0.07 & 0.52 & 0.40 & 1.25 & 61.5 & 100.0 \\
\rowcolor{rowDecision} lev & 83.4 & 56.7 & 0.18 & $-$0.01 & $-$0.04 & 0.14 & 0.08 & 1.55 & 58.2 & 100.0 \\
\midrule
\rowcolor{rowLLM} GPT-5.6-Luna & 91.2 & 77.7 & 0.78 & 0.87 & 0.68 & 0.89 & 0.89 & 0.48 & 93.3 & 100.0 \\
\rowcolor{rowLLM} Gemini-3.8-Flash & 78.9 & 70.2 & 0.83 & 0.84 & 0.64 & 0.89 & 0.86 & 0.70 & 84.6 & 100.0 \\
\rowcolor{rowLLM} Qwen-3.8-27B & 82.1 & 68.7 & 0.74 & 0.76 & 0.54 & 0.88 & 0.86 & 0.71 & 84.8 & 100.0 \\
\rowcolor{rowLLM} GPT-6-Luna & 90.1 & 64.5 & 0.65 & 0.60 & 0.34 & 0.75 & 0.69 & 0.88 & 80.1 & 100.0 \\
\rowcolor{rowLLM} Qwen-3.6-35B-A3B & 87.2 & 63.2 & 0.59 & 0.56 & 0.27 & 0.85 & 0.81 & 0.83 & 80.9 & 100.0 \\
\rowcolor{rowLLM} Claude-Haiku-5.5 & 87.6 & 52.0 & 0.56 & 0.80 & 0.51 & 0.83 & 0.80 & 0.69 & 89.1 & 100.0 \\
\rowcolor{rowLLM} Gemini-3.5-Flash-Lite & 77.7 & 58.0 & 0.67 & 0.78 & 0.15 & 0.83 & 0.79 & 0.99 & 72.3 & 100.0 \\
\rowcolor{rowLLM} Qwen-3.5-9B & 88.0 & 58.3 & 0.41 & 0.49 & 0.35 & 0.80 & 0.76 & 0.94 & 76.2 & 100.0 \\
\rowcolor{rowLLM} Gemma-4-26B-A4B & 65.5 & 62.4 & 0.64 & 0.74 & 0.23 & 0.76 & 0.73 & 1.26 & 59.8 & 100.0 \\
\rowcolor{rowLLM} Gemma-4-E2B & 81.4 & 47.5 & 0.28 & 0.57 & 0.16 & 0.58 & 0.64 & 1.12 & 69.0 & 100.0 \\
\rowcolor{rowLLM} Gemma-4-E4B & 87.1 & 53.0 & 0.06 & 0.30 & 0.22 & 0.65 & 0.70 & 1.03 & 72.8 & 100.0 \\
\rowcolor{rowLLM} Gemma-3-4B & 74.8 & 38.7 & 0.28 & 0.48 & 0.19 & 0.42 & 0.36 & 1.28 & 57.5 & 100.0 \\
\rowcolor{rowLLM} Qwen-3.5-2B & 77.1 & 37.8 & 0.05 & $-$0.04 & 0.04 & 0.20 & 0.25 & 1.97 & 40.8 & 100.0 \\
\bottomrule
\end{tabular}
\end{adjustbox}
\caption{{Agent workflow} judging against GPT-6.1-Sol on 8,600 conversations: raw agreement (\%) on the pass and task-completion verdicts, Spearman $\rho$ per rubric dimension, MAE and within-one agreement (W-1, \%) over the five dimensions, and percentage of conversations with a valid output.}
\label{tab:judge-skills}
\end{table*}

\newtcolorbox{promptbox}[1]{enhanced, colback=gray!4, colframe=gray!55, boxrule=0.5pt, arc=1.5pt,
  left=4pt, right=4pt, top=3pt, bottom=3pt, fonttitle=\bfseries\small, coltitle=black, colbacktitle=gray!18,
  title={#1}, fontupper=\footnotesize\ttfamily, before skip=6pt, after skip=8pt}

\section{\new{Per Task Sample Prompt}}
\label{app:prompts}
\new{Each evaluation instance consists of a context (e.g., conversation transcript), one or more questions, and the allowed answers for each question (Section~\ref{sec:setup}). Because the tasks come from production contact center workflows, the boxes below show generic templates rather than the exact prompts used in production. Decision models receive each question with its typed answer options; generative LLMs receive the same content together with the output instruction in the last box.}

\begin{promptbox}{Call Outcome}
CONTEXT\\
\{call transcript\}\\[3pt]
QUESTION\\
Which label best describes the outcome of this call?\\[3pt]
POSSIBLE ANSWERS\\
One of a fixed set of outcome labels with short descriptions (e.g., issue resolved, transferred).
\end{promptbox}

\begin{promptbox}{Purpose of Call}
CONTEXT\\
\{call transcript\}\\[3pt]
QUESTION\\
Which label best describes why the customer called?\\[3pt]
POSSIBLE ANSWERS\\
One of a fixed set of purpose labels (e.g., billing question).
\end{promptbox}

\begin{promptbox}{Quality Assurance}
CONTEXT\\
\{call transcript\}\\[3pt]
QUESTION (one per criterion)\\
Does the call satisfy \{service-quality criterion\}?\\[3pt]
POSSIBLE ANSWERS\\
Yes | No
\end{promptbox}

\begin{promptbox}{Procedural Step}
CONTEXT\\
\{call transcript\}\\[3pt]
QUESTION (one per step)\\
Did the agent carry out \{procedural step, e.g., verifying the caller's identity\}?\\[3pt]
POSSIBLE ANSWERS\\
Yes | No
\end{promptbox}

\begin{promptbox}{Intent}
CONTEXT\\
\{customer utterance\}\\[3pt]
QUESTION\\
Which configured intent best matches the customer's request?\\[3pt]
POSSIBLE ANSWERS (4--21 per sample)\\
\{intent\}: \{description\} | \ldots{} | no intent
\end{promptbox}

\begin{promptbox}{RAG Document}
CONTEXT\\
\{conversation\}; \{article 1\}, \{article 2\}, \{article 3\}\\[3pt]
QUESTION\\
Which article answers the customer's last question?\\[3pt]
POSSIBLE ANSWERS\\
1 | 2 | 3 | $-1$ (none relevant)
\end{promptbox}

\begin{promptbox}{Agent Workflow Judge}
CONTEXT\\
\{agent instructions and policies\}; \{conversation and tool-call log\}\\[3pt]
QUESTIONS\\
Was the customer's task completed?\\
Did the agent claim an action it did not perform?\\
Rate \{criterion\} from 1 to 5 (five criteria).\\[3pt]
POSSIBLE ANSWERS\\
success | failure | {partial} | appropriately transferred | appropriately out of scope;\\
yes | no;\quad 1--5
\end{promptbox}

\begin{promptbox}{Generated Resp.\ Judge: pointwise}
CONTEXT\\
\{generation instruction and call transcript\}; \{candidate response\}\\[3pt]
QUESTIONS\\
Rate the response from 1 to 5 for faithfulness, completeness, and instruction following.\\[3pt]
POSSIBLE ANSWERS (per question)\\
1 | 2 | 3 | 4 | 5
\end{promptbox}

\begin{promptbox}{Generated Resp.\ Judge: pairwise}
CONTEXT\\
\{generation instruction and call transcript\}; \{response A\}; \{response B\} (both orders)\\[3pt]
QUESTION\\
Which response is better?\\[3pt]
POSSIBLE ANSWERS\\
A | B | tie | both bad
\end{promptbox}

\begin{promptbox}{Output instruction for generative LLMs}
Return one JSON object that maps each question to one of its allowed answers.
\end{promptbox}

\end{document}